\pdfoutput=1

\documentclass{article}
\PassOptionsToPackage{numbers, compress}{natbib}
\usepackage[preprint]{neurips_2025}

\usepackage{wrapfig}
\usepackage{amssymb}
\usepackage{hyperref}
\usepackage{appendix}
\usepackage{listings}
\usepackage{booktabs}
\usepackage{multirow}
\usepackage{longtable}
\usepackage{array}
\usepackage{times}
\usepackage{latexsym}
\usepackage{float} 
\usepackage[T1]{fontenc}
\usepackage[utf8]{inputenc}
\usepackage{microtype}
\usepackage{inconsolata}
\usepackage{graphicx}
\usepackage[table]{xcolor} 
\usepackage{amsmath} 
\PassOptionsToPackage{numbers, compress}{natbib}
\usepackage{natbib}
\newcommand{\gptfour}{\mbox{GPT-4o}}
\newcommand{\geminitwo}{\mbox{Gemini 2.0}}

\title{A Cognitive Paradigm Approach to Probe the Perception-Reasoning Interface in VLMs} 

\author{%
Mohit Vaishnav\textsuperscript{\textnormal{1,2}}~~ 
Tanel Tammet\textsuperscript{\textnormal{1}} \\
   \textsuperscript{1} Applied Artificial Intelligence Group, Tallinn University of Technology, Estonia\\
   \textsuperscript{2} Kimova AI, Tallinn, Estonia \\ 
  \texttt{mohit.vaishnav@taltech.ee}
}

\begin{document}
\maketitle
\begin{abstract}

A fundamental challenge in artificial intelligence involves understanding the cognitive mechanisms underlying visual reasoning in sophisticated models like Vision-Language Models (VLMs). How do these models integrate visual perception with abstract thought, especially when reasoning across multiple images or requiring fine-grained compositional understanding? Drawing inspiration from cognitive science, this paper introduces a structured evaluation framework using diverse visual reasoning tasks – Bongard Problems (BPs) and Winoground – to dissect the perception-reasoning interface in VLMs. We propose three distinct evaluation paradigms, mirroring human problem-solving strategies: \textit{Direct Visual Rule Learning} (DVRL; holistic processing), \textit{Deductive Rule Learning} (DRL; rule extraction and application), and \textit{Componential Analysis} (CA; analytical decomposition via task-agnostic textual descriptions). These paradigms systematically vary cognitive load and probe processing stages. Notably, CA enables multi-image reasoning evaluation even for single-image architectures and isolates reasoning from perception by operating on textual descriptions. Applying this framework, we demonstrate that CA, leveraging powerful language models for reasoning over rich, independently generated descriptions, achieves new state-of-the-art (SOTA) performance on challenging benchmarks including Bongard-OpenWorld, Bongard-HOI, and Winoground. Ablation studies confirm reasoning improves significantly when perceptual challenges are mitigated, revealing a critical perception bottleneck. Our framework provides a valuable diagnostic tool and suggests that decoupling perception (via rich, task-agnostic description) from reasoning is a promising direction for robust and general visual intelligence.
\end{abstract}

\section{Introduction}

Human cognition adeptly integrates visual perception with abstract reasoning to navigate and understand the world \citep{kunda2020ai,lake2017building}. A central goal in AI is to imbue machines with similar visual intelligence \citep{cao2024visual}. While Vision-Language Models (VLMs) show remarkable progress \citep{zhao2023survey, radford2021learning}, the underlying mechanisms enabling complex visual reasoning – especially tasks demanding abstraction across multiple images or fine-grained compositional understanding – remain opaque. How do VLMs bridge pixels and concepts, and how do their strategies compare to human cognition?

To investigate these questions, we utilize two distinct classes of challenging visual reasoning tasks. First, Bongard Problems (BPs) \citep{bongard1968recognition}, a classic test requiring few-shot discovery of an abstract visual rule distinguishing positive and negative image sets. BPs demand perception, abstraction, and multi-image comparison, echoing human concept learning \citep{hofstadter1995fluid}. We use natural image variants, Bongard-OpenWorld \citep{wu2024bongardopenworld} and Bongard-HOI \citep{jiang2022bongard}. Second, we use Winoground \citep{thrush2022winoground}, which tests visio-linguistic compositional reasoning by requiring fine-grained differentiation between minimally contrastive image-caption pairs.

This paper introduces an evaluation framework explicitly designed to probe the cognitive processes underlying VLM performance on such challenging visual reasoning tasks, using natural image BPs \citep{wu2024bongardopenworld} as our primary testbed. Our core contribution is an \textbf{evaluation methodology} grounded in cognitive science principles. We propose three distinct paradigms that structure the BP task differently, mirroring established human problem-solving strategies:
\begin{enumerate}
    \item \textbf{Direct Visual Rule Learning (DVRL):} Simulates holistic or gist-based processing \citep{biederman87}, requiring the model to analyze all images simultaneously.
    \item \textbf{Deductive Rule Learning (DRL):} Mimics explicit, rule-based deduction \citep{rips1994psychology}, separating rule extraction from subsequent application.
    \item \textbf{Componential Analysis (CA):} Parallels analytical decomposition \citep{GLUCK2008193}, requiring models to reason over structured textual descriptions of individual images.
\end{enumerate}
This framework allows systematic analysis of VLM behavior under different cognitive demands and identify specific bottlenecks.

The CA paradigm offers unique advantages. It facilitates multi-image reasoning evaluation even for single-image VLM architectures. Crucially, by generating comprehensive, \textit{task-agnostic} image descriptions first and then performing reasoning solely over this text, CA allows us to \textit{disentangle perception from reasoning}. Furthermore, by substituting externally generated high-quality descriptions, we can isolate reasoning capabilities from the model's own perceptual limits and even evaluate text-only LLMs (Section \ref{ablation:description_quality}).

Applying this framework, we achieve \textit{new state-of-the-art (SOTA) results} on Bongard-OpenWorld, Bongard-HOI (on several splits), and Winoground using the CA paradigm, primarily by combining high-fidelity descriptions with powerful reasoning models (including \gptfour{}, \geminitwo{}, and specialized LLMs). This success across diverse tasks suggests the robustness and generality of decoupling perception into rich textual representations and leveraging strong language-based reasoning. Concurrently, our analysis reveals a significant \textit{perception bottleneck} in many open-source VLMs, whose performance drastically improves when their perceptual front-end is bypassed.

\textbf{Our contributions are thus:} (1) A novel, cognitively-inspired framework for the diagnostic evaluation of VLM visual reasoning. (2) Three evaluation paradigms enabling analysis of different reasoning strategies and facilitating multi-image task evaluation for diverse architectures. (3) A method (CA) for helping to disentangle perception from reasoning in VLMs and extending evaluation to LLMs. (4) Empirical results highlighting both the potential of advanced VLMs and a critical perception bottleneck in many current open-source models. (5) Evidence supporting the benefit of structured, multi-stage processing for VLM visual reasoning -- SOTA performance on multiple benchmarks (Bongard-OW, HOI, Winoground)

\section{Related Work}

Evaluating the burgeoning capabilities of Vision-Language Models (VLMs) necessitates diverse benchmarks. While foundational tasks like Visual Question Answering (VQA) \citep{antol2015vqa, gurari2018vizwiz} remain important, the focus has shifted towards complex reasoning \citep{zhang2024if, lu2022learn} and multi-modal, multi-image understanding, utilizing interleaved corpora \citep{zhu2024multimodal, laurenccon2024obelics} and dedicated benchmarks \citep{wu2023q, li2023fine, meng2024mmiu}. However, many evaluations rely heavily on linguistic context or do not specifically isolate the abstract visual concept formation and relational reasoning central to purely visual challenges like Bongard Problems (BPs).

Purely visual reasoning benchmarks provide a clearer window into perceptual and non-verbal reasoning abilities. BPs \citep{bongard1968recognition} are a canonical example, testing few-shot abstract rule discovery. Natural image variants like Bongard-HOI \citep{jiang2022bongard} and Bongard-OpenWorld \citep{wu2024bongardopenworld} enhance ecological validity, demanding comparison across complex images akin to human concept learning. Alongside other abstract reasoning tests like Raven's Progressive Matrices (RPMs) \citep{barrett2018measuring, zhang2019raven}, BPs provide a crucial testbed for advanced visual intelligence. Winoground \citep{thrush2022winoground} specifically targets visio-linguistic compositional understanding. Our contribution here an evaluation methodology designed to analyze \textit{how} models approach such cognitively demanding tasks.

Our evaluation framework is explicitly grounded in cognitive science perspectives on human problem-solving \citep{newell1972human}. The proposed paradigms reflect diverse cognitive strategies: \textit{Direct Visual Rule Learning (DVRL)} mirrors rapid holistic or similarity-based processing common in initial visual analysis \citep{biederman87, von2014neural}; \textit{Deductive Rule Learning (DRL)} reflects explicit, deliberative rule-based deduction \citep{rips1994psychology, knauff2010complex}; and \textit{Componential Analysis (CA)} parallels analytical decomposition where problems are broken into constituent features or components for systematic reasoning \citep{GLUCK2008193, hutchinson1992challenges}. Assessing VLMs through these distinct lenses provides a structured way to probe their internal processing characteristics relative to known cognitive modes.

Our approach connects to Chain-of-Thought (CoT) prompting methods \citep{wei2022chain, kojima2022large} that structure reasoning, including recent multimodal extensions \citep{zhang2023multimodal, chen2024visual} such as DDCoT \citep{zheng2023ddcot}, CoCoT \citep{zhang2024cocot}, and Compositional CoT \citep{mitra2024compositional}. While sharing the goal of step-wise processing, a key distinction lies in our Componential Analysis (CA) paradigm: whereas many multimodal CoT approaches generate intermediate representations often conditioned on the specific task context (e.g., descriptions tailored to a given caption \citep{zheng2023ddcot, zhang2024cocot}), CA deliberately generates comprehensive, task-independent textual image descriptions in a separate first stage, with subsequent reasoning operating only on this output. This decoupling appears key to the generalizability and SOTA performance we observe. Our paradigms act as diagnostic \textit{evaluation conditions} manipulating task structure, rather than solely performance-boosting prompts. Consequently, our paradigms function primarily as distinct, cognitively-motivated \textit{evaluation conditions} that manipulate the task structure itself (e.g., comparing holistic vs. deductive processing, separating perception from reasoning via CA) for diagnostic insights into VLM processes, rather than being prompting techniques aimed solely at maximizing task performance. 

\section{Models}

We selected a diverse suite of contemporary VLMs for evaluation, encompassing both leading closed-source systems and widely used open-source models of varying scales. This includes: \gptfour{} \citep{chatgpt}, \geminitwo{} \citep{gemini}, Pixtral-12B \citep{agrawal2024pixtral}, Llama-Vision-3.2 (11B, 90B) \citep{meta2024llama}, LLaVA (Llama-2 based; 7B, 13B, 34B) \citep{liu2023llava}, and LLaVA-Llama3-8B \citep{xtuner2025llava_llama3}. For ablation studies involving reasoning over text (Section \ref{ablation:description_quality}), we also included text-only LLMs (e.g., Phi-4 \citep{abdin2024phi};  Qwen2.5  \citep{qwen2.5}; Deepseek-r1 \citep{guo2025deepseek}; Gemma2 \citep{team2024gemma}). Model access was via official APIs or the Ollama framework \citep{ollama2025} running locally on NVIDIA GPUs (2080Ti, 3090, 6000 Ada). All evaluations employed few-shot prompting with zero temperature for deterministic output; no model fine-tuning was performed. Appendix \ref{app:models_exp} provides comprehensive details on models, configurations, etc.

\section{Dataset and Task}
\label{sec:dataset}

\begin{wrapfigure}{r}{0.5\textwidth}
\vspace{-1.cm}
\centering
\includegraphics[width=\linewidth]{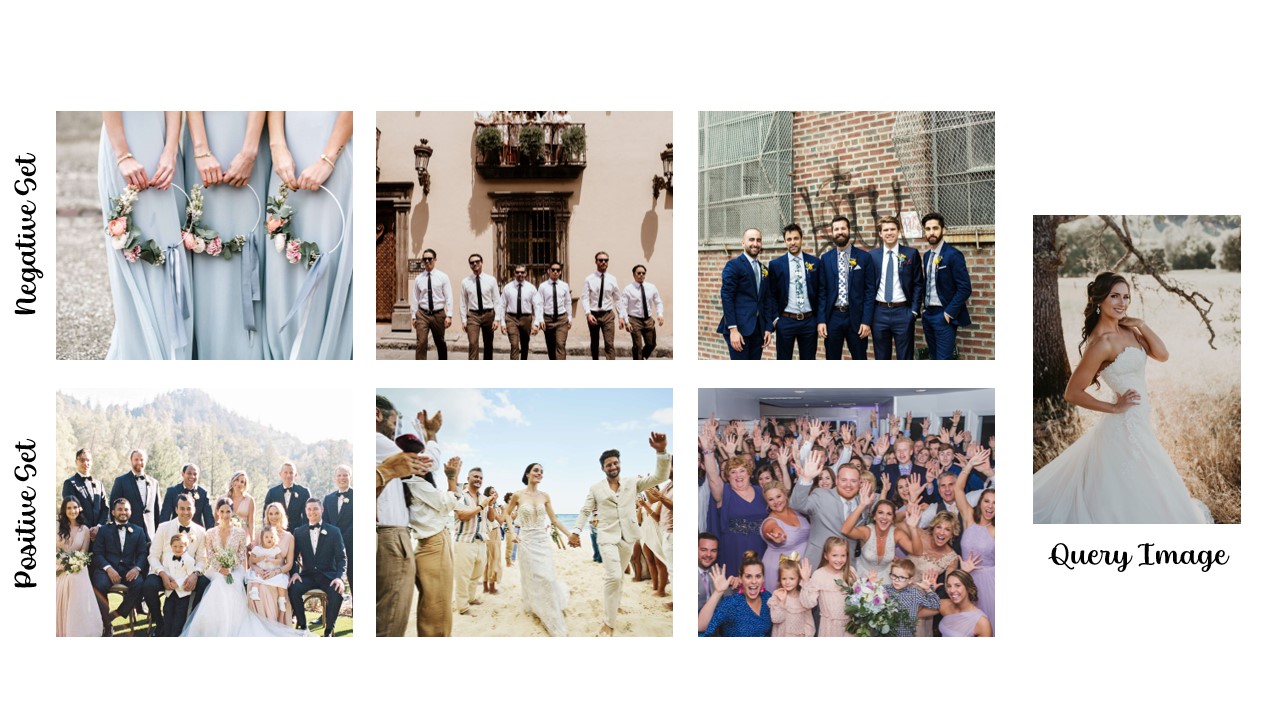} 
\vspace{-.6cm}
\caption{Example Bongard-OpenWorld task. \textit{Left}: Positive examples. \textit{Center}: Negative examples. \textit{Right}: Query. Rule: \textit{A group photo at a wedding reception}. Query is negative. (3 of 6 examples shown per set).}
\label{fig:bongard_example}
\end{wrapfigure}

Our primary testbed is the Bongard-OpenWorld dataset \citep{wu2024bongardopenworld}. We selected this benchmark for its use of natural, real-world images and its requirement for few-shot abstract reasoning based on commonsense visual concepts, aligning with the cognitive challenges inherent in BPs. From the full dataset, we constructed a balanced subset of 500 test cases (see Appendix \ref{app:dataset} for sampling details). Each case comprises 6 positive images exemplifying a rule, 6 negative images violating it, and 1 query image for classification. The dataset's grounding in real-world imagery and abstract rules provides a demanding test for VLM visual intelligence. An example task is shown in Figure \ref{fig:bongard_example}. For generalizability assessment, we also utilize the Bongard-HOI \citep{jiang2022bongard} and Bongard Logo \citep{nie2020bongard} datasets (Section \ref{result:generalizability}).

We use additional two distinct visual reasoning benchmarks:\\
\begin{itemize}
    \item \textbf{Bongard-HOI (BP-HOI)} \citep{jiang2022bongard}: Focuses on rules based on human-object interactions in natural images. Used 4 standard splits (N=100 each).\\
    \item \textbf{Winoground} \citep{thrush2022winoground}: Assesses fine-grained compositional visio-linguistic reasoning via image-caption matching (N=400 samples used).
\end{itemize}

These diverse tasks probe abstraction (BPs) and compositionality (Winoground), providing a comprehensive testbed.

\section{Cognitively-Inspired Evaluation Paradigms}
\label{sec:paradigms}

We evaluate VLMs using three paradigms designed to probe different facets of visual reasoning, inspired by human cognitive strategies. All paradigms require the model to output a structured response including analysis, the derived rule, query description, and classification (\texttt{positive}/\texttt{negative}). Figure \ref{fig:cognitive_paradigm} provides a schematic overview. Specific prompts are detailed in Appendix \ref{app:prompts}.

\begin{figure*}[htbp] 
\centering
\includegraphics[width=\linewidth]{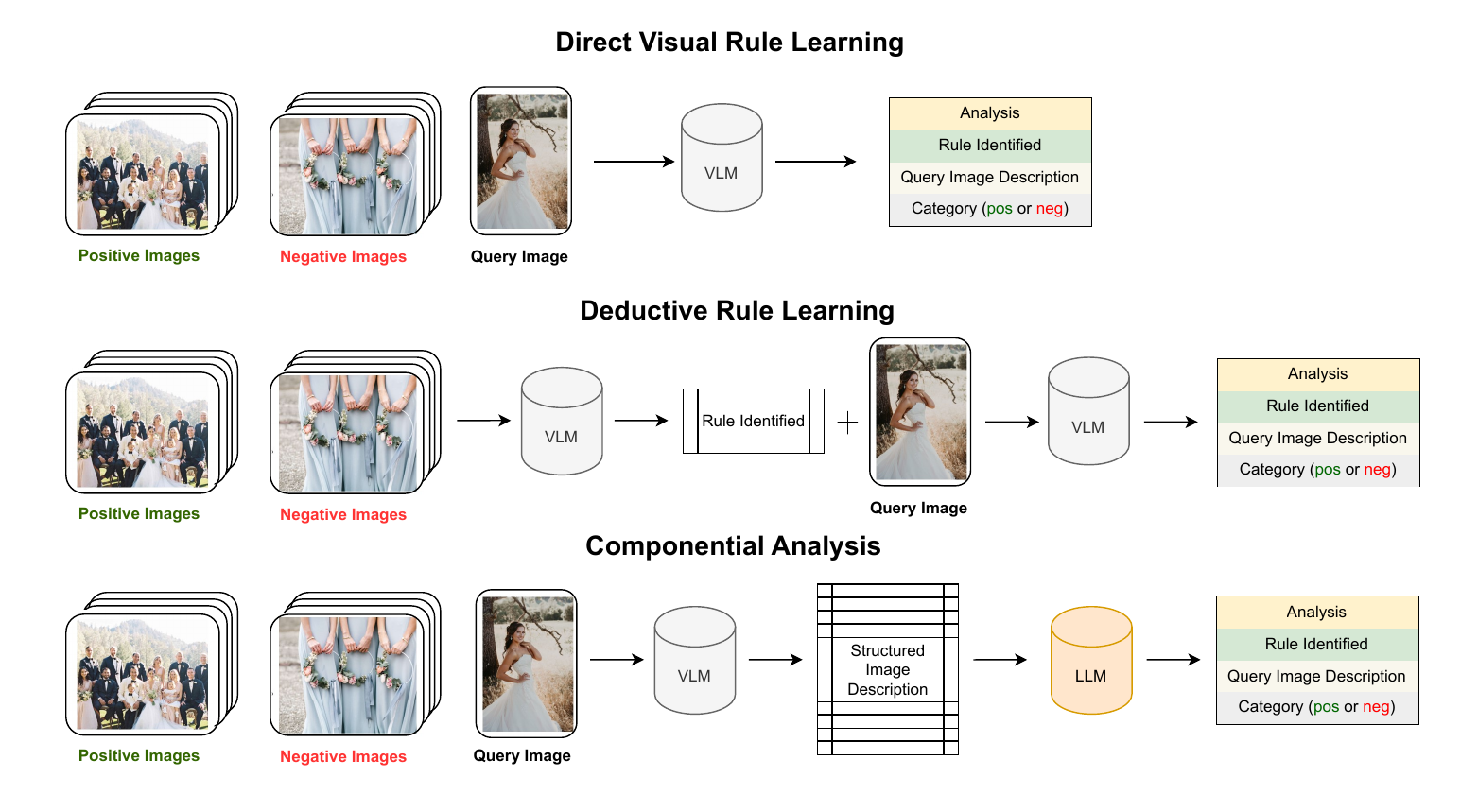} 
\caption{\textbf{Cognitively-Inspired Evaluation Paradigms.} \textbf{DVRL} (Direct Visual Rule Learning): Concurrent processing of all images, mimicking holistic perception. Requires multi-image input capability. \textbf{DRL} (Deductive Rule Learning): Two-stage process separating rule extraction from application, mimicking explicit deduction. \textbf{CA} (Componential Analysis): Multi-stage process involving individual image description followed by reasoning over text, mimicking analytical decomposition and enabling perception-reasoning separation.}
\label{fig:cognitive_paradigm}
\end{figure*}

\subsection{Direct Visual Rule Learning (DVRL)}
\label{exp:direct}
This paradigm assesses holistic reasoning by presenting all 13 images (6 positive, 6 negative, 1 query) simultaneously to the VLM. It demands the model integrate information across the entire set to identify the rule and classify the query in one step. This mirrors the human ability to quickly grasp the `gist' of a visual scene or problem. Due to requiring simultaneous multi-image input, only models like \geminitwo{} and \gptfour{} were tested under this paradigm.

\subsection{Deductive Rule Learning (DRL)}
\label{exp:deductive}
Mimicking deliberative, rule-based deduction, DRL involves two stages:
\begin{enumerate}
    \item \textbf{Rule Extraction:} The VLM analyzes the 12 context images (positive/negative sets) to identify and concisely summarize (max 20 words) the distinguishing rule.
    \item \textbf{Rule Application:} The VLM receives the previously generated rule summary and the query image, classifying the query based solely on the provided rule.
\end{enumerate}
This separation allows examining the fidelity of both rule formation and rule application processes.

\subsection{Componential Analysis (CA)}
\label{exp:componential_analysis}
Reflecting analytical problem decomposition, CA proceeds in stages based on textual representations:
\begin{enumerate}
    \item \textbf{Image Description:} The VLM generates a detailed, structured JSON description for each of the 13 images \textit{individually}.
    \item \textbf{Text-Based Reasoning:} The VLM receives the collection of 13 JSON descriptions (labeled positive/negative/query) and performs rule extraction and query classification based \textit{only} on this textual input.
\end{enumerate}
This paradigm is crucial as it (a) allows evaluating models lacking direct multi-image input, (b) enables assessing reasoning largely independent of perceptual errors by potentially using externally generated descriptions (see Section \ref{ablation:description_quality}), and (c) facilitates evaluation of text-only LLMs on visual reasoning tasks.

\section{Results and Analysis}

This section details the performance of the evaluated VLMs, beginning with the primary Bongard-OpenWorld benchmark and then examining generalizability.

\subsection{Performance on Bongard-OpenWorld}

\begin{table*}[htbp] 
\centering
\setlength{\tabcolsep}{2pt} 
\begin{tabular}{lccccccccc}
\toprule
\multirow{2}{*}{\textbf{Model}} & \multicolumn{3}{c}{\textbf{DVRL (\%)}} & \multicolumn{3}{c}{\textbf{DRL (\%)}} & \multicolumn{3}{c}{\textbf{CA (\%)}} \\
\cmidrule(lr){2-4} \cmidrule(lr){5-7} \cmidrule(lr){8-10}
 & \texttt{neg} & \texttt{pos} & Overall & \texttt{neg} & \texttt{pos} & Overall & \texttt{neg} & \texttt{pos} & Overall \\
\midrule
\rowcolor{gray!20} 
\gptfour{} \citep{chatgpt} & 66.4 & 93.6 & 80.0 & 82.8 & 93.2 & 88.0 & 92.8 & 92.8 & \textbf{92.8} \\
\rowcolor{gray!20} 
\geminitwo{} \citep{gemini} & 80.5 & 83.9 & 82.2 & 85.7 & 87.9 & 86.8 & 90.8 & 96.4 & \textbf{93.6} \\
Pixtral-12B & - & - & - & - & - & - & 88.8 & 85.6 & 87.2 \\
Llama-Vision-11B \citep{meta2024llama} & - & - & - & - & - & - & 42.8 & 64.0 & 53.4 \\
Llama-Vision-90B \citep{meta2024llama} & - & - & - & - & - & - & 58.7 & 51.5 & 55.1 \\ 
Llava-7B \citep{liu2023llava} & - & - & - & - & - & - & 65.2 & 67.2 & 66.2 \\
Llava-Llama3-8B \citep{xtuner2025llava_llama3}& - & - & - & - & - & - & 93.0 & 13.1 & 53.2 \\
\midrule
\rowcolor{gray!20} 
Human Average \citep{wu2024bongardopenworld} & - & - & - & - & - & - & \multicolumn{2}{c}{(across samples)} & 91.0 \\
\bottomrule
\end{tabular}
\caption{Classification accuracy (\%) across evaluation paradigms on the Bongard-OpenWorld subset. Paradigms abbreviated: DVRL, DRL, CA. Dashes (-) indicate non-applicability due to model input limitations.}
\label{tab:merged_results}
\end{table*}

Table \ref{tab:merged_results} presents the core results on our 500-sample Bongard-OpenWorld subset.

Under \textbf{Direct Visual Rule Learning (DVRL)}, applicable only to \gptfour{} and \geminitwo{}, performance was strong but below optimal (\geminitwo{}: 82.2\%, \gptfour{}: 80.0\%), suggesting limitations in purely holistic, simultaneous multi-image reasoning for this complex task.

Performance improved markedly under \textbf{Deductive Rule Learning (DRL)} for both models (\gptfour{}: 88.0\%, \geminitwo{}: 86.8\%). The explicit separation of rule extraction and application stages appears beneficial, aligning with the idea that breaking down complex cognitive tasks can improve performance.

\textbf{Componential Analysis (CA)}, reasoning over textual descriptions, yielded the highest accuracies for the top models (\gptfour{}: 92.8\%, \geminitwo{}: 93.6\%), surpassing the reported human average \citep{wu2024bongardopenworld} for this benchmark. This underscores their powerful text-based reasoning ability when the visual input is effectively encoded into language. Pixtral-12B also achieved strong CA performance (87.2\%). However, a significant gap emerged with other open-source models. Models like Llama-Vision and LLaVA variants exhibited much lower CA accuracy, often with dramatic imbalances between positive and negative sample performance (e.g., Llava-Llama3-8B: 53.2\% overall, heavily biased towards negative samples). This pattern strongly suggests that the bottleneck for these models is not necessarily the abstract reasoning itself, but rather the fidelity of their internal \textit{visual perception} and subsequent translation into usable representations (here, text descriptions).

The consistent trend of accuracy increasing from DVRL through DRL to CA for \gptfour{} and \geminitwo{} further reinforces the value of structured reasoning and, particularly, the effectiveness of the component-based textual reasoning approach for this task when perception is adequate.

\subsection{Performance on Bongard HOI}
\label{result:generalizability}

\begin{wraptable}{r}{7.5cm}
\vspace{-.4cm}
\centering
\setlength{\tabcolsep}{2pt} 
\begin{tabular}{llccccc}
\toprule
\textbf{Model} & \textbf{Paradigm} & \textbf{sosa} & \textbf{soua} & \textbf{uosa} & \textbf{uoua} & \textbf{Avg} \\
\midrule
\multirow{3}{*}{\geminitwo{}}
    & DVRL & 50  & 54  & 49  & 50 & 50.8 \\
    & DRL  & 63  & 62  & 55  & 65 & 61.3  \\
    & CA   & 77  & 74  & \cellcolor{gray!20}70  & \cellcolor{gray!20}77 & 74.5 \\
\midrule
\multirow{3}{*}{\gptfour{}}
    & DVRL & 68  & 75  & 61  & 70 & 68.5 \\
    & DRL  & 73  & 77  & 64  & 73 & 71.8 \\
    & CA   & \cellcolor{gray!20}83  & \cellcolor{gray!20}83  & 66  & \cellcolor{gray!20}77 & 77.3 \\
\midrule
Human Avg. & -- & 87.2 & 90.0 & 93.6 & 94.9 & 91.4 \\
\bottomrule
\end{tabular}
\caption{Performance (\%) on Bongard-HOI splits across paradigms. Human average taken from \citep{jiang2022bongard} \protect\small{\textbf{Splits:} sosa: seen\_obj\_seen\_act, soua: seen\_obj\_unseen\_act, uosa: unseen\_obj\_seen\_act, uoua: unseen\_obj\_unseen\_act. Human average from cited source.}}
\label{tab:hoi_results}
\end{wraptable}

On Bongard-HOI, we evaluated \gptfour{} and \geminitwo{} across the four standard test splits (\texttt{sosa}, \texttt{soua}, \texttt{uosa}, \texttt{uoua}; N=100 each). The results, shown in Table \ref{tab:hoi_results}, largely replicated the trends observed on Bongard-OpenWorld. Performance systematically improved with increased paradigm structure (DVRL < DRL < CA) for both models. CA yielded the highest accuracies (65-83\% range), generally surpassing DRL (64-77\%) and DVRL (61-75\%). This consistency validates the framework's applicability and the benefit of structured evaluation across different complex natural image reasoning datasets. Notably, model performance is lower overall on HOI compared to OpenWorld, suggesting HOI presents distinct challenges, potentially in identifying subtle interaction-based rules. The human average scores reported in \citep{jiang2022bongard} are high (87-95\%), indicating a significant gap remains even for top models on this dataset.

\subsection{Performance on Winoground}
\label{sec:winoground_results}

\begin{table}[ht]
\centering
\setlength{\tabcolsep}{2pt}
\begin{tabular}{lccc}
\toprule
\textbf{Model} & \textbf{Text Score} & \textbf{Image Score} & \textbf{Group Score} \\
\midrule
\rowcolor{gray!20}
\textbf{\gptfour{}} \citep{chatgpt} & \textbf{75.50} & \textbf{58.50} & \textbf{52.00} \\
\geminitwo{} \citep{gemini} & 71.00 & 48.75 & 42.00 \\ 
\midrule
Llama3.3-70B (\citep{dubey2024llama}) & 68.25 & 49.25 & 41.75 \\ 
Qwen2.5-32B (\citep{qwen2.5}) & 67.00 & 46.25 & 40.00 \\ 
Phi-4-14B (\citep{abdin2024phi}) & 65.25 & 46.00 & 37.75 \\ 
Qwen2.5-14B (\citep{qwen2.5}) & 59.25 & 34.50 & 27.25 \\ 
\midrule
MMICL + CoCoT \citep{zhang2024cocot} & 
64.25 & 52.5 & 50.75 \\
\bottomrule
\end{tabular}
\vspace{0.5em}
\caption{State-of-the-art performance on the \textbf{Winoground benchmark} achieved using our \textbf{Componential Analysis (CA)} paradigm. Scores reported are the standard Winoground metrics: \textit{Text Score} (correct caption selection per image description), \textit{Image Score} (correct image description selection per caption), and \textit{Group Score} (all selections correct per sample), averaged over 400 samples.}
\label{tab:winoground_sota}
\end{table}

We applied our CA paradigm to the distinct challenge of Winoground \citep{thrush2022winoground}, which requires fine-grained compositional reasoning to match images and captions. Stage 1 generated task-agnostic descriptions for each image using \geminitwo{}, and Stage 2 used various LLMs (acting as the reasoning engine) to perform the matching based \textit{only} on the descriptions and captions (see Appendix \ref{app:winoground_scoring} for metric details).

As shown in Table \ref{tab:winoground_sota}, this approach achieves \textit{new state-of-the-art results} on Winoground across all three metrics (Text, Image, Group scores). Using \gptfour{} as the reasoning engine in CA yields scores of Text: 75.5\%, Image: 58.5\%, Group: 52.0\%. This significantly surpasses previously reported SOTA. Even using \geminitwo{} or strong open LLMs like Llama3-70B or Qwen2.5-32B within the CA framework produces highly competitive or SOTA results.

This successful application to Winoground demonstrates that our CA evaluation strategy is not limited to rule-discovery tasks like Bongard problems but can be effectively adapted to probe compositional understanding, further supporting the generalizability and diagnostic potential of our cognitively-inspired framework. This success is significant because our CA approach generates image descriptions \textit{independently} of the captions or the matching task itself. Unlike prior methods that often condition visual analysis on textual context \citep{zheng2023ddcot, zhang2024cocot}, our task-agnostic description generation followed by powerful text-based reasoning proves highly effective for fine-grained compositionality. This suggests that decoupling rich perception (into text) from robust linguistic reasoning offers a generalizable and high-performing strategy for complex visio-linguistic tasks. 



\section{Ablation Studies: Isolating Perception and Reasoning}
\label{sec:ablation}

To further investigate the interplay between visual perception, rule representation, and reasoning, we conducted targeted ablation studies. Both studies presented below serve to underscore the critical role of the initial representation derived from visual input – whether it's applying a rule \textit{to} a perceived query image (Section \ref{ablation:rule_application}) or reasoning \textit{from} perceived context images (Section \ref{ablation:description_quality}).

\subsection{Rule Application Fidelity}
\label{ablation:rule_application}

\begin{wraptable}{r}{6cm}
\vspace{-.4cm}
\centering
\setlength{\tabcolsep}{4pt} 
\begin{tabular}{lccc}
\toprule
\multirow{2}{*}{\textbf{Model}} & \multicolumn{3}{c}{\textbf{ Accuracy (\%)}} \\
\cmidrule(lr){2-4}
 & Overall & \texttt{neg} & \texttt{pos} \\
\midrule
LLaVA-7B & 72.0 & 52.0 & 92.0 \\
Llama-vision-11B & 68.2 & 38.0 & 98.4 \\
Pixtral-12B & 88.0 & 86.8 & 89.2 \\ 
LLaVA-13B & 70.0 & 58.8 & 81.2 \\
LLaVA-34B & 74.8 & 52.8 & 96.8 \\
Llama-vision-90B & 74.2 & 59.6 & 88.8 \\
\bottomrule
\end{tabular}
\caption{\textbf{Rule Application Accuracy:} Performance when classifying query images based on externally provided rules.}
\vspace{-.4cm}
\label{tab:rule_eval_results}
\end{wraptable}

How well can models apply an abstract rule once it's formulated? To isolate rule application from rule extraction, we provided models with high-quality rule summaries (generated by \gptfour{}) and the query image, tasking them solely with classification based on the given rule. This tests the model's ability to ground the symbolic rule in the visual input of the query image.

Table \ref{tab:rule_eval_results} shows performance for several open-source models under this condition. While models like Pixtral-12B demonstrate relatively strong and balanced rule application (88.0\% overall), others exhibit a confirmation bias, performing well on positive queries but poorly on negatives (e.g., Llama-Vision-11B: 98.4\% positive vs. 38.0\% negative accuracy). This suggests difficulty in reliably identifying when an image \textit{fails} to meet the rule criteria, even when the rule is provided explicitly. Comparing this to Table \ref{tab:merged_results}, the generally higher scores here than in CA (where models generate their own descriptions) support the idea that rule application itself is less challenging for these models than the initial perception/description phase.

\subsection{Impact of Description Quality on Reasoning}
\label{ablation:description_quality}
\begin{wraptable}{r}{.5\linewidth}
\vspace{-1.4cm}
\setlength{\tabcolsep}{2pt} 

\centering
\begin{tabular}{lc}
\toprule
\textbf{Model} & \textbf{Accuracy} \\
\midrule
Llava:7b \citep{liu2023llava} & 80.56 \\
Llava:34b \citep{liu2023llava} & 81.56 \\
Llama3.2-vision:11b \citep{meta2024llama} & 84.17 \\
Llama3.2-vision:90b \citep{meta2024llama} & 90.98 \\
\midrule
Deepseek-r1:14b \citep{guo2025deepseek} & 87.98 \\
Gemma2:27b \citep{team2024gemma} & 88.98 \\
Qwen2.5:7b \citep{qwen2.5} & 90.38 \\
Phi4:14b \citep{abdin2024phi} & 91.98 \\
Qwen2.5:32b \citep{qwen2.5} & 92.79 \\
Qwen2.5:14b \citep{qwen2.5} & 92.99 \\
\bottomrule
\end{tabular}
\caption{\textbf{Impact of High-Quality Descriptions}: Accuracy using Componential Analysis (Stage 2 reasoning) with image descriptions generated externally by \gptfour{}. Includes VLMs and LLMs models respectively separated by line.}
\label{tab:description_quality_impact} 
\vspace{-.7cm}
\end{wraptable}

Complementing the previous ablation, we investigated how reasoning performance changes when the initial perceptual stage (description generation) is standardized using a high-fidelity source. We generated descriptions for all context and query images using \gptfour{} and then used these descriptions as input to the reasoning stage (Stage 2) of the Componential Analysis paradigm for various target models, including weaker VLMs and even text-only LLMs.

The results, presented in Table \ref{tab:description_quality_impact}, were revealing. Providing high-quality descriptions dramatically improved the reasoning accuracy of VLMs that struggled when using their own descriptions. Llama-Vision-11B, for example, improved from 53.4\% (Table \ref{tab:merged_results}) to 84.17\%, and Llama-Vision-90B from 55.1\% to 90.98\%. This provides strong evidence that the reasoning capabilities of these models are significantly underestimated by end-to-end evaluations; their primary limitation lies in generating accurate perceptual representations. Further illustrating this sensitivity to description source quality, Table~\ref{tab:components_comparison} in the Appendix details a comparison using components generated by Pixtral-12B.

Remarkably, this approach also enabled text-only LLMs to perform the visual reasoning task effectively. Models such as Phi-4 (14B) achieved 91.98\% accuracy, surpassing the average human baseline, while several Qwen models also exceeded 90\%. This demonstrates that: (1) High-quality textual descriptions can serve as effective surrogates for visual input, enabling modality transfer for reasoning tasks. (2) The CA paradigm, particularly when coupled with controlled descriptive input, serves as a powerful tool for isolating and evaluating the core symbolic reasoning abilities of both VLMs and LLMs, independent of their integrated perceptual systems. These findings strongly reinforce the conclusion that improving visual perception is paramount for enhancing end-to-end visual reasoning in many current models.

\subsection{Semantic Similarity Analysis}
\label{ablation:semantic} 
Semantic similarity analysis during DRL (Table \ref{tab:semantic_similarity}) confirmed that derived rules generally aligned well with query descriptions, particularly for positive samples. The relatively high similarity for negative samples highlights the challenge of the dataset's near-miss counterexamples.

\subsection{Qualitative Error Analysis}
\label{sec:error_analysis_main}
Examining samples misclassified by both top models (\gptfour{}, \geminitwo{}) under CA revealed recurring error patterns (details in Appendix \ref{app:error_analysis}, Table \ref{tab:tasks}). Frequent issues involved over-generalizing rules, missing critical objects/properties present in positive examples, focusing on spurious correlations, or failing to consistently apply derived rules. These qualitative examples underscore that even highly capable models exhibit fragility in nuanced visual detail processing and robust symbolic rule manipulation.

\section{Discussion}

This research leveraged a cognitively-inspired framework to dissect the visual reasoning mechanisms of VLMs. By evaluating performance across paradigms mirroring human strategies (holistic, deductive, analytical), we moved beyond accuracy to probe \textit{how} models process visuo-conceptual information. A central finding is the critical role, and frequent limitation, of \textit{visual perception}. While advanced VLMs possess powerful reasoning capabilities, unlocked effectively by the Componential Analysis (CA) paradigm when using high-fidelity textual descriptions, many models are bottlenecked by their inability to reliably extract and represent relevant visual information.

Our findings highlight the critical interdependence of perception and reasoning. While human cognition often involves rapid perceptual grouping (`gist' perception \citep{biederman87, Li2002}) followed by more focused attention and deliberation \citep{Posner1980, kahneman2011thinking}, our results suggest many current VLMs struggle significantly with the initial perceptual stage. The strong performance of advanced models like \gptfour{} and \geminitwo{} under the CA paradigm – where visual complexity is reduced to textual descriptions – indicates powerful downstream reasoning capabilities. However, the dramatic performance drop for many open-source models in this same paradigm (when using their own descriptions) points squarely to a \textit{perception bottleneck}.

The success of the CA paradigm, achieving SOTA on diverse benchmarks like Bongard-OpenWorld, Bongard-HOI, and notably Winoground, is particularly insightful. Its strength appears to lie in \textit{decoupling perception from reasoning via task-agnostic description generation}. Unlike end-to-end models or context-dependent CoT approaches that might conflate perceptual errors with reasoning failures or overfit to linguistic cues, CA first aims to build a comprehensive, independent representation of each visual input in text. This rich textual world model then allows powerful LLMs (whether multi-modal or text-only) to apply their sophisticated reasoning abilities effectively, leading to robust and generalizable performance across different reasoning types (abstraction in BPs, compositionality in Winoground). This suggests modularity, where specialized perception modules generate rich symbolic representations for general-purpose reasoning engines, might be a highly effective architecture.

The ablation studies reinforce this conclusion. Providing external rules (Section \ref{ablation:rule_application}) isolates rule application fidelity, while providing high-quality external descriptions (Section \ref{ablation:description_quality}) isolates reasoning from perception. The results clearly show that reasoning performance is significantly enhanced when the perceptual input is reliable. This suggests that for many systems, failures attributed to ``reasoning'' may actually stem from noisy or inaccurate initial visual representations. From a cognitive perspective, this is akin to reasoning failures caused by misperception rather than flawed logic.

The progressive performance improvement observed across paradigms (DVRL < DRL < CA for top models on natural images) suggests that structuring the reasoning process, particularly by separating distinct cognitive stages like rule extraction and application, or perception and reasoning, benefits current VLM architectures. This resonates with findings on Chain-of-Thought prompting \citep{wei2022chain} but our staged paradigms offer a different way to structure and diagnose the process, particularly relevant for multi-modal tasks involving complex visual comparisons.

Furthermore, the CA paradigm demonstrates a viable approach for evaluating multi-image reasoning tasks even with single-image input models. By converting images to text, it allows models without inherent multi-image capacity to engage with tasks like BPs. It also successfully \textit{bridged the modality gap}, enabling LLMs to perform sophisticated visual reasoning when provided with rich descriptions, achieving high performance. This highlights the potential of symbolic representations derived from vision as a powerful interface for general reasoning engines.


Overall, our cognitively-inspired evaluation framework provides valuable diagnostic tools. It helps pinpoint specific weaknesses, such as the perception bottleneck identified in open-source models, and reveals the latent reasoning potential unlocked when perceptual challenges are mitigated.

\section{Limitations}
\label{sec:limitation}
This work, while providing insights, has limitations. Our primary analysis focuses on multi image visual reasoning benchmark; further validation on other single image visual reasoning benchmarks would further strengthen the generality of our findings regarding the perception bottleneck and the utility of the paradigms. The effectiveness of Componential Analysis hinges on the quality and nature of textual descriptions; it may be less suited for tasks involving non-compositional or purely geometric/topological rules that are difficult to articulate linguistically. Our analysis of computational demands was superficial; a deeper investigation into the efficiency trade-offs of these multi-stage paradigms is warranted. Finally, while inspired by human cognition, our paradigms are simplified models, and VLM internal processes may differ significantly from their human counterparts. 

\section{Conclusion}

This paper introduced a cognitively-inspired evaluation framework to dissect the perception-reasoning interface in VLMs using diverse visual reasoning tasks. Through three paradigms (DVRL, DRL, CA), we systematically analyzed VLM processing. Our key contribution is the Componential Analysis (CA) approach, which decouples perception into task-agnostic textual descriptions, enabling robust reasoning by powerful language models. This method achieves \textit{new state-of-the-art results} on challenging benchmarks requiring different reasoning styles: Bongard-OpenWorld (abstraction), Bongard-HOI (interaction), and Winoground (compositionality). This success highlights the power of rich symbolic representations derived from vision. Concurrently, our framework diagnoses a critical \textit{perception bottleneck} limiting many current VLMs, whose reasoning abilities are significantly unlocked when provided with high-fidelity descriptions (confirmed via ablations and LLM evaluations). Our work provides valuable diagnostic tools and suggests that enhancing perceptual fidelity and exploring decoupled perception-reasoning architectures are crucial steps towards achieving general, robust visual intelligence in AI.

\bibliography{custom} 

\clearpage
\appendix
\section*{Appendix}
\label{sec:appendix}

\renewcommand{\thefigure}{A.\arabic{figure}}
\renewcommand{\thesection}{A.\arabic{section}}
\renewcommand{\thetable}{A.\arabic{table}}
\setcounter{figure}{0}
\setcounter{section}{0}
\setcounter{table}{0}


\section{Broader Relevance}
\label{app:broader}
This study offers insights with broader implications for developing more robust and human-like AI systems. Our cognitively-inspired evaluation paradigms provide valuable tools for assessing and understanding the strengths and limitations of Vision-Language Models (VLMs) on complex visual reasoning tasks. The insights gained extend beyond Bongard problems, contributing to the development of VLMs capable of advanced reasoning in real-world applications. Our key finding regarding the visual processing bottleneck in many models has significant implications for future research aimed at bridging the performance gap and unlocking the full potential of accessible models. The demonstration of high performance by advanced VLMs underscores the potential for sophisticated visual understanding, reinforcing the importance of architectures integrating robust perception and reasoning. Finally, our comparative evaluation contributes to discussions about AI accessibility and transparency, identifying specific areas for improvement and paving the way for more reliable AI.

\section{Attention and Memory in Visual Reasoning}
\label{app:attenmemory}
While our study primarily focuses on the interplay between perception and reasoning, the roles of attention and memory are also implicitly present in our paradigms. The DVRL paradigm likely engages VLM ``visual attention'' mechanisms \citep{bahdanau2016neuralmachinetranslationjointly} to identify salient features across the image set, akin to human holistic processing \citep{biederman87, Li2002}. DRL relies on the model's ability to ``memorize'' the extracted rule, involving processes related to working memory \citep{Baddeley2012} and internal representation storage \citep{Squire1992}. Although not directly measured, their involvement is inherent. Future work could explore these aspects more explicitly, perhaps via attention map analysis \citep{NIPS2017_3f5ee243} or probing memory representations \citep{vaishnav2023gamr}.

\section{Dataset Details}
\label{app:dataset}
\subsection{Bongard OpenWorld Dataset}
We utilize a subset of 500 test cases from the Bongard OpenWorld dataset \citep{wu2024bongardopenworld}. The full dataset contains 1001 samples, each with 7 positive and 7 negative real-world images distinguished by a ``commonsense'' rule. Our evaluation set was created by taking the first 250 samples and generating two test cases from each (one positive query, one negative query), resulting in 500 balanced test cases. Specific sample IDs used will be released.

\subsubsection{Commonsense Value Categories}
\label{app:commonsense_cats}
Table \ref{tab:concepts_appendix} summarizes the rule categories.

\begin{table*}[htbp] 
\centering
\begin{tabular}{cll}
\toprule 
\textbf{ID} & \textbf{Concept Category} & \textbf{Example} \\
\midrule 
0 & Anything else & Animals are running. \\
1 & Human-Object Interaction (HOI) & A person playing the guitar. \\
2 & Taste / Nutrition / Food & A plate of high-calorie food. \\
3 & Color / Material / Shape & A wooden floor in the living room. \\
4 & Functionality / Status / Affordance & An animal capable of flying in the tree. \\
5 & And / Or / Not & A man without beard. \\
6 & Factual Knowledge & A building in US capital. \\
7 & Meta Class & Felidae animals. \\
8 & Relationship & A bench near trees. \\
9 & Unusual Observations & Refraction of light on a glass cup. \\
\bottomrule 
\end{tabular}
\caption{Commonsense ID Categories and Examples in Bongard OpenWorld dataset \citep{wu2024bongardopenworld}.} 
\label{tab:concepts_appendix} 
\end{table*}

\subsubsection{Commonsense Value Distribution in Our Subset}
\label{app:commonsense_dist}
Table \ref{tab:commonsense_dist_appendix} shows the distribution in our subset. Category `0' is predominant.

\begin{table}[htbp] 
\centering
\begin{tabular}{crr}
\toprule 
\textbf{ID} & \textbf{Count} & \textbf{Percentage (\%)} \\
\midrule 
0 & 365 & 73.0 \\
3 & 36 & 7.2 \\
9 & 26 & 5.2 \\
1 & 15 & 3.0 \\
2 & 14 & 2.8 \\
4 & 12 & 2.4 \\
5 & 10 & 2.0 \\
6 & 10 & 2.0 \\
8 & 8 & 1.6 \\
7 & 4 & 0.8 \\
\midrule 
\textbf{Total} & \textbf{500} & \textbf{100.0} \\
\bottomrule 
\end{tabular}
\vspace{0.5em}
\caption{Distribution of Commonsense ID Categories in the 500 Bongard-OpenWorld test cases used in our evaluation subset.} 
\label{tab:commonsense_dist_appendix} 
\end{table}

\subsection{Bongard-HOI Dataset} 
\label{app:dataset_hoi} 

To assess generalizability on natural images with a different reasoning focus (human-object interactions), we used the Bongard-HOI dataset \citep{jiang2022bongard}. We evaluated performance on its four standard test splits, defined by object/action novelty:
\begin{itemize}
    \itemsep0em
    \item \texttt{sosa}: seen object, seen action
    \item \texttt{soua}: seen object, unseen action
    \item \texttt{uosa}: unseen object, seen action
    \item \texttt{uoua}: unseen object, unseen action
\end{itemize}
The original splits vary significantly in size and balance (e.g., \texttt{sosa}: 200 pos/200 neg queries; \texttt{soua}: 2236 pos/1348 neg; \texttt{uosa}: 660 pos/660 neg; \texttt{uoua}: 695 pos/695 neg). For consistent cross-split evaluation in this work, we created balanced subsets by sampling 100 test cases from each of the four splits, ensuring an equal distribution of 50 positive and 50 negative query images per split. This resulted in a total evaluation set of 400 samples for Bongard-HOI (100 per split), used for the results reported in Table~\ref{tab:hoi_results}.

\subsection{Winoground Dataset} 
\label{app:dataset_wino} 

To test performance on fine-grained visio-linguistic compositional reasoning, we utilized the Winoground dataset \citep{thrush2022winoground}. This dataset comprises 400 samples specifically designed to challenge compositional understanding. Each sample contains a pair of minimally contrastive images ($I_0, I_1$) and a corresponding pair of minimally contrastive captions ($C_0, C_1$), requiring models to correctly match image $I_0$ to caption $C_0$ and image $I_1$ to caption $C_1$. We used all 400 samples provided in the standard dataset release for our Winoground evaluations reported in Section~\ref{sec:winoground_results} and Table~\ref{tab:winoground_sota}. 

\subsection{Dataset Availability}
\label{app:dataset_availability}
Bongard OpenWorld: \url{https://rujiewu.github.io/Bongard-OpenWorld.github.io/}. \\
Bongard-HOI: \url{https://github.com/NVlabs/Bongard-HOI/blob/master/assets/dataset.md}. \\
Winoground: \url{https://huggingface.co/datasets/facebook/winoground}

Details on the specific subsets and samples used in our evaluations will be released upon publication.

\section{Model and Experiment Details}
\label{app:models_exp}
\subsection{Model Details}
\label{app:models}
\textbf{VLMs:} \gptfour{}; \geminitwo{}; Pixtral-12B; Llama-Vision-3.2 (11B, 90B); LLaVA (Llama-2 based; 7B, 13B, 34B); LLaVA-Llama3-8B.
\textbf{Text-Only LLMs (for Ablation \ref{ablation:description_quality}):} Phi-4 (14B) \citep{abdin2024phi};  Qwen2.5 (7B, 14B, 32B) \citep{qwen2.5}; Deepseek-r1 (32B, 70B) \citep{guo2025deepseek}; Gemma2 (27B) \citep{team2024gemma}.

\subsection{Experiment Configuration}
\label{app:config}
\begin{itemize}
    \item \textbf{Access:} APIs for closed models; Ollama for open models.
    \item \textbf{Input:} Base64 images in prompts (see Appendix~\ref{app:prompts}).
    \item \textbf{Image Handling:} API defaults or max 1024px (Ollama). Multi-image calls for DVRL where supported.
    \item \textbf{Decoding:} Temperature 0.
    \item \textbf{Fine-tuning:} None.
    \item \textbf{Hardware:} NVIDIA GPUs (2080Ti, 3090, 6000 Ada).
\end{itemize}

\subsection{Evaluation Metrics}
\label{app:eval}
\begin{itemize}
    \item \textbf{Classification Accuracy:} Primary metric (\% correct).
    \item \textbf{Semantic Similarity:} Cosine similarity of OpenAI embeddings (`text-embedding-3-large`) between descriptions/rules. Inspired by \citep{risch-etal-2021-semantic}.
\end{itemize}

\subsection{Winoground Score Calculation using Componential Analysis}
\label{app:winoground_scoring}

This section defines the calculation of Winoground \citep{thrush2022winoground} scores (\texttt{text\_score}, \texttt{image\_score}, \texttt{group\_score}) within our Componential Analysis (CA) paradigm (Section \ref{result:generalizability}).

In the standard Winoground task, a sample $i$ consists of two images $I_{0,i}, I_{1,i}$ and two captions $C_{0,i}, C_{1,i}$, where $(C_{0,i}, I_{0,i})$ and $(C_{1,i}, I_{1,i})$ are the ground truth correct pairs. Models are typically evaluated based on a scoring function $s(C, I)$ indicating the match between a caption and an image.

In our CA paradigm, the reasoning model (Stage 2) does not access images $I_{0,i}, I_{1,i}$ directly. Instead, it operates on textual image descriptions $D_{0,i}, D_{1,i}$ generated in Stage 1. The model is prompted to make explicit choices about the best match between descriptions and captions. Let $Choice_C(D_k, \{C_0, C_1\})$ denote the caption ($C_0$ or $C_1$) chosen by the model as the best match for description $D_k$. Similarly, let $Choice_D(C_k, \{D_0, D_1\})$ denote the description ($D_0$ or $D_1$) chosen for caption $C_k$.

The scores for each sample $i$ in the dataset $W$ (where $N = |W| = 400$) are calculated as follows:

\textbf{1. Text Score ($f_{CA}$):} This measures if the correct caption is selected for each image description. We use an indicator function $\mathbb{I}[\cdot]$ which is 1 if the condition inside is true, and 0 otherwise.
\begin{equation}
\label{eq:text_score_ca}
f_{CA}(i) = \mathbb{I} \Big[ \begin{gathered} Choice_C(D_{0,i}, \{C_{0,i}, C_{1,i}\}) = C_{0,i} \\ and \\ Choice_C(D_{1,i}, \{C_{0,i}, C_{1,i}\}) = C_{1,i} \end{gathered} \Big]
\end{equation}
This score is 1 only if the model correctly identifies the caption for \textit{both} description $D_{0,i}$ and description $D_{1,i}$.

\textbf{2. Image Score ($g_{CA}$):} This measures if the correct image description is selected for each caption.
\begin{equation}
\label{eq:image_score_ca}
g_{CA}(i) = \mathbb{I} \Big[ \begin{gathered} Choice_D(C_{0,i}, \{D_{0,i}, D_{1,i}\}) = D_{0,i} \\ and \\ Choice_D(C_{1,i}, \{D_{0,i}, D_{1,i}\}) = D_{1,i} \end{gathered} \Big]
\end{equation}
This score is 1 only if the model correctly identifies the description for \textit{both} caption $C_{0,i}$ and caption $C_{1,i}$.

\textbf{3. Group Score ($h_{CA}$):} This requires all associations within the sample to be correct.
\begin{equation}
\label{eq:group_score_ca}
h_{CA}(i) = f_{CA}(i) \land g_{CA}(i)
\end{equation}
Equivalently, $h_{CA}(i) = 1$ if and only if $f_{CA}(i) = 1$ and $g_{CA}(i) = 1$.

\lstset{language=Python, 
        basicstyle=\ttfamily\small, 
        keywordstyle=\color{green}, 
        stringstyle=\color{red}, 
        commentstyle=\color{blue}, 
        showstringspaces=false}

\lstset{language=Python, 
        basicstyle=\ttfamily\small, 
        keywordstyle=\color{green}, 
        stringstyle=\color{red}, 
        commentstyle=\color{blue}, 
        showstringspaces=false,
        breaklines=true,          
        breakatwhitespace=true    
}

\section{Model Prompts}
\label{app:prompts}
\subsection{Direct Visual Rule Learning}
\label{app:holistic_prompt}
The prompt used for the Direct Visual Rule Learning paradigm is designed to elicit a holistic analysis of the provided images, encouraging the model to identify a distinguishing rule and apply it to the query image.  The prompt emphasizes the distinction between positive ($cat\_2$) and negative ($cat\_1$) examples and guides the model to provide a structured output containing its analysis, the identified rule, details about the query image, and the final classification.

\begin{lstlisting}
    
def visual_concept_test_prompt(m, n): 
    """
    Generates a visual analysis prompt.
    
    Args:
        m (int): Number of positive samples.
        n (int): Number of negative samples.
    
    Returns:
        str: The formatted prompt string.
    """
    return f"""
    You are provided with {m + n + 1} images: the first {m} samples are `cat_2`, the next {n} samples are `cat_1`, and the last image is the `query image`. 
    Analyze the common characteristics or patterns found in the `cat_2` samples (positive samples: following 1 common rule) that distinctly separate them from the `cat_1` samples (negative samples: it might not follow any possible rule). 
    Your task is to:
    
    1. Determine the rule or criterion that distinguishes the `cat_2` samples from the `cat_1` ones.
    2. Analyse the `query image` (last image).
    3. Provide your conclusion for the `query image` if it can be categorized as either `cat_1` or `cat_2` based on the analysis and the rule.

    Ensure that the output is clear, well-formatted, and free of unnecessary explanations. 
    Omit the ``` tags at the beginning and end of the page. The format of your output should be as follows:

    - **Analysis**: (Your analysis here)
    - **Rule**: (The distinguishing rule here)
    - **Query Image**: (Query image details)
    - **Conclusion**: (cat_1 or cat_2)
    """
\end{lstlisting}

\subsection{Deductive Rule Learning}
The Deductive Rule Learning paradigm employs a two-stage prompting strategy.  The first stage focuses on rule extraction from positive and negative examples, while the second stage applies the extracted rule to classify a query image.  The prompts for each stage are detailed below.

\subsubsection{First-Stage Prompt (Rule Extraction)}
\label{app:rule_extraction_prompt}

This prompt guides the model to identify and summarize a distinguishing rule based on provided positive and negative examples.  It emphasizes conciseness in the rule summary.

\begin{lstlisting}
    def visual_concept_prompt(m, n):
    try:
        if m < 0 or n < 0:
            raise ValueError(f"Invalid input: m and n must be non-negative. Received m={m}, n={n}.")
        
        if m > 0 and n > 0:
            prompt = f"""
                You are provided with {m + n} images: the first {m} samples are cat_2, the next {n} samples are cat_1. Analyze the common characteristics or patterns found in the cat_2 samples (positive samples: following 1 common rule) that distinctly separate them from the cat_1 samples (negative samples: it might not follow any possible rule). 
                Your task is to provide the rules that defines cat_2 samples. At the end, write "summary" of the rule identified in less than 20 words.
                Ensure that the output is clear, well-formatted, and free of unnecessary explanations. Omit the ``` tags at the beginning and end of the page.
                """
        if n == 0:
            prompt = f"""
                You are provided with {m} images: {m} samples are cat_2. Analyze the common characteristics or patterns found in the cat_2 samples (positive samples: following 1 common rule) that distinctly separate them from negative samples which might not follow any possible rule. 
                Your task is to provide the rules that defines cat_2 samples. At the end, write "summary" of the rule identified in less than 20 words.
                Ensure that the output is clear, well-formatted, and free of unnecessary explanations. Omit the ``` tags at the beginning and end of the page.
                """
        return prompt
    
    except ValueError as e:
        print(f"Error: {e}")
        raise
\end{lstlisting}

\subsubsection{Second-Stage Prompt (Rule Application)}
\label{app:rule_application_prompt}

This prompt presents the previously extracted rule summary and a query image, prompting the model to classify the image based on the rule.  It reinforces the Bongard problem context and requests a structured output.

\begin{lstlisting}
# Define the visual analysis prompt
def visual_concept_test_prompt(m, n, summary):
    return f"""
    We are working with Bongard dataset where there are {m} image in the cat_2 and {n} images in the cat_1. Summary of the common characteristics or patterns found in the cat_2 samples (positive samples: following 1 common rule) that distinctly separate them from the cat_1 samples (negative samples: it might not follow any possible rule) is as follows: \n {summary}. 
    
    Your task is to ponder over the rule and provide your conclusion for the `query image` if it can be categorized as either "cat_1" or "cat_2".

    Ensure that the output is clear, well-formatted, and free of unnecessary explanations. 
    Omit the ``` tags at the beginning and end of the page. The format of your output should be as follows:

    - **Analysis**: (Your analysis here)
    - **Rule**: (The distinguishing rule here)
    - **Query Image**: (Query image details)
    - **Conclusion**: (cat_1 or cat_2)
    """
\end{lstlisting}

\subsection{Componential Analysis}
\label{app:componential_prompts}

The Componential Analysis paradigm also uses a two-stage prompting strategy. The first stage generates detailed image descriptions, while the second stage derives a rule from these descriptions and applies it to a query image.  The specific prompts for each stage are presented below.

\subsubsection{First-Stage Prompt (Image Description Generation)}
\label{app:image_description_prompt}

This prompt instructs the model to generate a comprehensive, hierarchical description of a given image in JSON format.  It guides the model to cover various aspects of the image, from scene and objects to activities and contextual elements, facilitating detailed comparative analysis in the subsequent stage.

\begin{lstlisting}
# Define the visual analysis prompt
def visual_concept_prompt(): 
    """
    Generates a visual analysis prompt.
    
    Args:
    
    Returns:
        str: The formatted prompt string.
    """
    return """ 
            Carefully examine the provided image and identify all possible visual elements, organizing them into a detailed hierarchical structure. Start with broad categories and progress to more specific subcategories. This should cover everything visible in the image, ensuring no detail is overlooked. Structure your findings in a JSON format to enable easy comparison and synthesis of data from other images. This will help discern patterns, contexts, and rules valuable for identifying or understanding query images.

            Your hierarchy might encompass the following elements:

            1. **Scene/Environment**: Description of the overall setting depicted, such as urban, natural, indoor, or outdoor scenes.
            2. **Objects**: Define distinct items or entities present in the scene.
            - **Living Beings**: Animals, humans, or other biological entities.
                - Species or classification (e.g., dog, bird, human).
                - Characteristics (e.g., color, posture, movement).
            - **Inanimate Objects**: Both synthetic and natural elements.
                - Categories (e.g., vehicle, building, trees).
                - Properties (e.g., color, size, material, shape).
            3. **Activities**: Observable actions or interactions involving any objects or beings.
            - Specific descriptions of actions (e.g., walking, flying).
            - Participants involved in these actions.
            4. **Contextual Elements**: Environmental conditions and time markers, such as time of day or weather.
            - Detailed characteristics (e.g., cloudy, night, winter).
            5. **Visual Patterns**: Prominent colors, textures, and patterns that are visually significant.
            6. **Emotional Undertones**: Any emotional presence or expressions evident in the image.
            7. **Textual Information**: Any visible text within the image, including what it says and its visual style.
            8. **Summary**: A concise narrative summarizing the overall content and context of the image.

            Ensure that every aspect from the image is represented under these categories. The information should be presented in the following JSON format:

            {
            "Scene": {
                "Description": "..."
            },
            "Objects": {
                "Living Beings": [...],
                "Inanimate Objects": [...]
            },
            "Activities": [...],
            "Contextual Elements": {
                "Time of Day": "...",
                "Weather": "..."
            },
            "Visual Patterns": {
                "Dominant Colors": [...],
                "Textures": [...]
            },
            "Emotional Undertones": "..."
            "Textual Information": "..."
            "Summary": "..."
            }
            Ensure that the JSON output is clear, well-formatted, and free of unnecessary explanations. Omit the ```json tags at the beginning and end of the page.
            """
\end{lstlisting}

\subsubsection{Second-Stage Prompt (Rule Derivation Instruction)}
\label{app:system_eval_prompt}

This prompt guides the model to analyze the JSON descriptions generated in the first stage, derive a distinguishing rule, and apply it to classify a query image.  It emphasizes the use of the provided JSON format and requests a structured output.

\begin{lstlisting}
def user_eval_prompt(all_image_specs, m, n):
    return f"""
        We are working with the Bongard dataset, which contains {m} images in cat_2 (positive samples) and {n} images in cat_1 (negative samples). These categories are defined as follows:
        - Cat_2: Positive samples that follow a single common rule.
        - Cat_1: Negative samples that may not follow any specific rule.

        The image descriptions for the positive samples, negative samples, and the test image are provided in JSON format. Analyze the common patterns or characteristics in the cat_2 samples that distinguish them from cat_1 samples.

        Your task is to:
        1. Derive the rule that defines the cat_2 samples.
        2. Apply this rule to categorize the test image.

        Here are the image descriptions:

        ### Positive Samples (cat_2):
        {all_image_specs[:m]}

        ### Negative Samples (cat_1):
        {all_image_specs[m:m+n]}

        ### Test Image:
        {all_image_specs[-1]}

        Provide your output in the following format:

        - **Analysis**: (Your analysis here)
        - **Rule**: (The distinguishing rule here)
        - **Test Image**: (Test image details)
        - **Conclusion**: (cat_1 or cat_2)
        """

\end{lstlisting}

\section{Results and extended analysis}
\subsection{Performance on Bongard Openworld}
\begin{table}[htbp] 
\centering
\setlength{\tabcolsep}{4pt} 
\begin{tabular}{lcccc}
\toprule
& \multicolumn{2}{c}{\textbf{\geminitwo{}}} & \multicolumn{2}{c}{\textbf{\gptfour{}}} \\
\textbf{Category} & \textbf{Mean} & \textbf{Std Dev} & \textbf{Mean} & \textbf{Std Dev} \\ 
\midrule
\texttt{Positive} & 0.915 & 0.02 & 0.902 & 0.02 \\ 
\texttt{Negative} & 0.868 & 0.02 & 0.866 & 0.02 \\ 
\bottomrule
\end{tabular}
\vspace{0.5em}
\caption{Semantic Similarity (Cosine) between query descriptions and rules derived during Deductive Rule Learning.}
\label{tab:semantic_similarity} 
\end{table}

\subsection{Performance on Bongard-HOI}
\label{app:hoi_results}
(Refer to Table \ref{tab:hoi_results} in main text)


\subsection{Winoground Performance Context}
\label{app:winoground_context}

\begin{table}[htbp]
\centering
\begin{tabular}{lccc}
\toprule
\textbf{Model / Strategy} & \textbf{Text} & \textbf{Image} & \textbf{Group} \\
\midrule
Gemini (Baseline) & 30.75 & 26.00 & 25.00 \\
Gemini + DDCoT  & 45.00 & 25.00 & 23.75 \\
Gemini + CCoT   & 22.50 & 33.00 & 20.75 \\
Gemini + CoCoT  & 40.00 & 32.50 & 27.75 \\
\midrule
\textbf{\geminitwo{} + CA (Ours)} & \textbf{71.91} & \textbf{48.71} & \textbf{42.01} \\
\bottomrule
\end{tabular}
\vspace{0.5em}
\caption{Performance comparison on Winoground (400 samples). CA refers to our Componential Analysis paradigm. Other results use Gemini Pro Vision with different prompting strategies.}
\label{tab:winoground_comparison}
\end{table}

To contextualize the performance of our Componential Analysis (CA) paradigm applied to \geminitwo{} on Winoground (reported in Section~\ref{result:generalizability}), we also ran evaluations using Gemini Pro Vision with several prompting strategies. Table~\ref{tab:winoground_comparison} shows these comparative results on the 400-sample Winoground set used. While advanced CoT methods like DDCoT and CoCoT improve over the baseline for Gemini Pro Vision, the CA paradigm applied to \geminitwo{} achieves competitive scores, particularly on the text metric, demonstrating its effectiveness.

\subsection{Comparison of Description Sources (Pixtral-12B vs. GPT-4o)}
\label{app:component_comparison} 

The results, detailed in Table~\ref{tab:components_comparison}, consistently show that using image components described by \gptfour{} yielded higher downstream reasoning accuracy compared to using components described by Pixtral-12B across all tested reasoning models. While both description sources enabled strong performance, the advantage conferred by \gptfour{}'s descriptions (ranging from approximately 2\% to over 11\% improvement depending on the reasoning model) further underscores the critical dependence of reasoning outcomes on the fidelity, richness, and potentially the alignment of the initial perceptual descriptions with the concepts required by the reasoning task. This reinforces the significance of the VLM's front-end visual processing and description capabilities as a key factor influencing overall visual reasoning performance.

\begin{table}[htbp]
\centering
\begin{tabular}{@{}lcc@{}}
\toprule
 & \multicolumn{2}{c}{\textbf{Components (\%)}} \\
\cmidrule(lr){2-3}
\textbf{Model} & \textbf{Pixtral-12B} & \textbf{\gptfour{}} \\
\midrule
Deepseek-R1-14B      & 83.21 & 87.98 \\
Llama3.2-vision-90B  & 89.05 & 90.98 \\
Phi-4-14B            & 86.86 & 91.98 \\
Qwen2.5-14B          & 90.51 & 92.99 \\
LLaVA-7B             & 68.61 & 80.56 \\
Llama3.2-vision-11B  & 80.29 & 84.17 \\
LLaVA-34B            & 79.56 & 81.56 \\
Phi-3-14B            & 84.67 & 86.97 \\
\bottomrule
\end{tabular}
\vspace{0.5em}
\caption{Performance comparison using Componential Analysis (Stage 2) with image descriptions generated by either Pixtral-12B or \gptfour{}. Evaluated across various reasoning models.}
\label{tab:components_comparison}
\end{table}

\subsection{Componential Analysis Results by Commonsense Category}
\label{app:commonsense_results_detailed} 
\begin{table}[htbp] 
\centering
\begin{tabular}{c >{\raggedright\arraybackslash}p{4.5cm} cc} 
\toprule 
\textbf{ID} & \textbf{Concept Category} & \textbf{\gptfour{} (\%)} & \textbf{\geminitwo{} (\%)} \\ 
\midrule 
0 & Anything else & 92.88 & 94.23 \\
1 & Human-Object Interaction (HOI) & 86.67 & 92.86 \\
2 & Taste / Nutrition / Food & 100.00 & 85.71 \\
3 & Color / Material / Shape & 88.89 & 91.67 \\
4 & Functionality / Status / Affordance & 100.00 & 100.00 \\ 
5 & And / Or / Not & 90.00 & 80.00 \\
6 & Factual Knowledge & 90.00 & 90.00 \\
7 & Meta Class & 100.00 & 100.00 \\
8 & Relationship & 100.00 & 100.00 \\
9 & Unusual Observations & 92.31 & 92.31 \\
\bottomrule 
\end{tabular}
\vspace{0.5em}
\caption{Overall accuracy (\%) of \gptfour{} and \geminitwo{} on the Bongard-OpenWorld test set using Componential Analysis, broken down by Commonsense ID category. Performance variations highlight differing model strengths on specific concept types.} 
\label{tab:commonsense} 
\end{table}
Analysis of \gptfour{} and \geminitwo{} performance in CA across commonsense categories (Appendix Table \ref{tab:commonsense}) showed generally strong performance, indicating robustness to varied conceptual rules. Minor variations suggested potential differences in handling specific types of context or attributes, possibly reflecting training data nuances.

\subsection{Impact of CoT-like Structure}
\label{app:cot_impact}
(Refer to Table \ref{tab:cot_results_appendix} below)

\begin{table}[h!] 
\centering
\begin{tabular}{lccc}
\toprule
\multirow{2}{*}{\textbf{Prompt Type}} & \multicolumn{3}{c}{\textbf{Accuracy (\%)}} \\
\cmidrule(lr){2-4}
 & Overall & \texttt{neg} & \texttt{pos} \\
\midrule
Minimal (No CoT) & 61.6 & 39.2 & 84.0 \\
Structured (CoT-like) & 80.0 & 66.4 & 93.6 \\
\bottomrule
\end{tabular}
\vspace{0.5em}
\caption{Impact of Structured Prompting on DVRL accuracy (GPT-4o).}
\label{tab:cot_results_appendix}
\end{table}

\subsection{Detailed Error Analysis Examples}
\label{app:error_analysis}
(Refer to Table \ref{tab:tasks} below)

\begin{longtable}{|c|c|>{\centering\arraybackslash}m{3cm}|>{\raggedright\arraybackslash}m{7cm}|} 
\hline
\textbf{No.} & \textbf{Test ID} & \textbf{Caption (Rule)} & \textbf{Reason for Error (Based on GPT-4o o/p)} \\ \hline 
\endfirsthead

\hline
\textbf{No.} & \textbf{Test ID} & \textbf{Caption (Rule)} & \textbf{Reason for Error (Based on GPT-4o Output)} \\ \hline
\endhead

\hline
\multicolumn{4}{|r|}{{Continued on next page}} \\ \hline
\endfoot

\hline
\endlastfoot

1 & 0021\_neg\_0 & Cars on the city streets at night & Weak reasoning (similarity): Rule requires vehicles, test image (painting) lacks them explicitly, though context implies city. \\ \hline
2 & 0014\_neg\_0 & A person playing a guitar. & Rule extraction error: Rule too general (e.g., ``person with instrument''), misses specific object (guitar) mentioned in analysis. \\ \hline
3 & 0033\_neg\_0 & A bicycle is placed in the corner & Rule extraction error: Misses key property (in a corner / specific placement context). Test image (collage) lacks this context. \\ \hline
4 & 0037\_neg\_0 & The girl has long and thin braids on her head. & Rule extraction error: Rule too general (e.g., ``girl with braids''), misses specific property (long and thin). \\ \hline
5 & 0076\_pos\_0 & Various kinds of rings & Rule extraction error: Rule misses specific object (ring), focuses on property (intricate design) absent in query. \\ \hline
6 & 0076\_neg\_0 & Various kinds of rings & Rule extraction error: Rule misses specific object (ring), too general. \\ \hline
7 & 0082\_neg\_0 & Live coral on the sea floor. & Weak reasoning (similarity): Rule identifies `coral', but test image description fails to mention it. Perceptual description error. \\ \hline
8 & 0084\_neg\_0 & A wooden fence surrounding a grassy field. & Rule extraction error: Rule misses specific object (grass), uses broader term (greenery). Test image has greenery but not clearly grass. \\ \hline
9 & 0112\_neg\_0 & A wooden floor in the living room. & Rule extraction error: Misses key objects (living room, floor), focuses only on `wooden' and general `indoor'. \\ \hline
10 & 0117\_neg\_0 & Colorful ribbons. & Rule extraction error: Rule too general, misses specific object (ribbons). \\ \hline
11 & 0122\_neg\_0 & A satellite view of Earth. & Rule extraction error: Misses specific viewpoint (top-down satellite), uses more general `aerial'. \\ \hline
12 & 0136\_pos\_0 & Spectator seats view in the stadium. & Weak reasoning/Rule Application error: Rule mentions ``sports or spectators'', query image description lacks both, leading to incorrect negative classification despite being stadium seats. \\ \hline
13 & 0213\_neg\_0 & Checkerboard pattern fabrics & Rule extraction error: Misses specific object context (fabric), although pattern is identified. \\ \hline
14 & 0234\_neg\_0 & A beautiful stone sculpture & Rule extraction error: Focuses on wrong property (`prominent' obelisk) instead of the intended rule property (`tall' obelisk). \\ \hline
15 & 0247\_pos\_0 & Small river filled with reeds & Rule extraction error: Misses key object (reeds), while focusing on negative constraints (no industrial presence) which are weakly present. \\ \hline \hline
\caption{Error Analysis: Examples of Bongard-OpenWorld cases misclassified by both GPT-4o and Gemini 2.0 in Componential Analysis. Captions indicate the ground truth rule \citep{wu2024bongardopenworld}. Reasoning based on analyzing GPT-4o's generated analysis, rule, and query description.}
\label{tab:tasks}
\end{longtable}

\end{document}